\pdfoutput=1

\documentclass[11pt]{article}

\usepackage{naacl2021}

\usepackage{times}
\usepackage{latexsym}

\usepackage[T1]{fontenc}

\usepackage[utf8]{inputenc}

\usepackage{microtype}

\usepackage{tabularx}
\usepackage{xspace}
\usepackage{multirow}

\usepackage{subcaption}
\usepackage{tabularx}
\usepackage{booktabs}
\usepackage{csvsimple}
\usepackage{enumitem}
\usepackage{soul}

\setlist{noitemsep}

\newcommand{\taskname}{\textsc{SCIVER}\xspace}
\newcommand{\numTeams}{11\xspace}
\newcommand{\numSubmissions}{14\xspace}
\newcommand{\improvement}{23\xspace}
\newcommand{\numClaims}{1,409\xspace}
\newcommand{\numTrain}{809\xspace}
\newcommand{\numDev}{300\xspace}
\newcommand{\numTest}{300\xspace}
\newcommand{\numPapers}{5,183\xspace}
\newcommand{\numParamsRobertaLarge}{355M\xspace}

\newcommand{\vertserini}{\textsc{Vert5Erini}\xspace}
\newcommand{\paragraphJoint}{\textsc{ParagraphJoint}\xspace}

\newcommand{\qmul}{\textsc{QMUL-SDS}\xspace}
\newcommand{\scikgat}{\textsc{SciKGAT}\xspace}
\newcommand{\fever}{\textsc{FEVER}\xspace}
\newcommand{\verisci}{\textsc{VeriSci}\xspace}

\newcommand{\nei}{\textsc{NEI}\xspace}
\newcommand{\support}{\textsc{Support}\xspace}
\newcommand{\refute}{\textsc{Refute}\xspace}
\newcommand{\supports}{\textsc{Supports}\xspace}
\newcommand{\refutes}{\textsc{Refutes}\xspace}

\newcommand{\scifact}{\textsc{SciFact}\xspace}

\definecolor{lightblue}{RGB}{212, 235, 255}
\definecolor{salmon}{RGB}{255, 164, 168}
\definecolor{lightgreen}{RGB}{177, 231, 171}
\definecolor{lightyellow}{RGB}{255, 255, 148}
\sethlcolor{lightblue}

\newcolumntype{Y}{>{\centering\arraybackslash}X}
\newcolumntype{s}{>{\hsize=.5\hsize}X}
\newcolumntype{L}[1]{>{\raggedright\let\newline\\\arraybackslash\hspace{0pt}}m{#1}}

\title{Overview and Insights from the \taskname Shared Task \\ on Scientific Claim Verification}

\author{David Wadden \\
  University of Washington \\
  \texttt{dwadden@cs.washington.edu} \\ \And
  Kyle Lo \\
  Allen Institute for AI \\
  \texttt{kylel@allenai.org} \\}

\begin{document}
\maketitle

\begin{abstract}
We present an overview of the \taskname shared task, presented at the 2nd Scholarly Document Processing (SDP) workshop at NAACL 2021. In this shared task, systems were provided a scientific claim and a corpus of research abstracts, and asked to identify which articles \support or \refute the claim as well as provide evidentiary sentences justifying those labels. \numTeams teams made a total of \numSubmissions submissions to the shared task leaderboard, leading to an improvement of more than +\improvement F1 on the primary task evaluation metric. In addition to surveying the participating systems, we provide several insights into modeling approaches to support continued progress and future research on the important and challenging task of scientific claim verification.
\end{abstract}

\section{Introduction}

Due to both rapid growth in scientific publications and the proliferation of mis- and dis-information online, there is a growing need for automated systems that can assist humans in assessing the veracity of scientific claims with evidence found in the research literature. For the \taskname shared task, systems are given a claim about a scientific entity and a corpus of abstracts from peer-reviewed research articles, and are asked to identify the articles in the corpus that \support or \refute the claim; each prediction is required to be accompanied with evidentiary sentences, or rationales, from the abstract that justify the labeling decision.

This task poses various challenges to systems.  For example, entailment relationships between claims and evidences found in scientific writing are often complex, and understanding them to arrive at a correct \support or \refute labeling decision may require reasoning about numerical quantities, increases and decreases in measurements, or causal chains. Furthermore, since annotations require scientific expertise, training data for this task are scarce. As a result, systems must employ techniques to overcome the lack of training data, such as domain adaptation or transfer learning.

Here, we report the results on the \taskname shared task. A total of \numTeams teams made \numSubmissions submissions to the shared task leaderboard, leading to a collective improvement of +\improvement F1 on the primary evaluation metric, compared to previous baselines. The remainder of this report is organized as follows: In \S \ref{sec:task}, we describe the shared task setup, including the choice of dataset, task definition, and evaluation metrics. In \S \ref{sec:systems}, we provide an overview of the systems submitted for the task, and highlight unique features of individual systems. In \S \ref{sec:results}, we present the results from the shared task evaluation. Finally, in \S \ref{sec:insights}, we identify several insights into modeling approaches demonstrated by participating systems.

\begin{table*}[t]
  \footnotesize
  \centering

  \begin{tabular}{lll}
    \toprule
    System Name           & Team affiliations                                  & Associated paper                       \\
    \midrule
    VerT5erini            & University of Waterloo                        & \citet{Pradeep2020ScientificCV} \\
    ParagraphJoint        & UT Dallas / Chan Zuckerburg Initiative / UCLA & \citet{Li2021APM}               \\
    Law \& Econ           & ETH Zurich                                    & \citet{Stammbach2020eFEVEREA}   \\
    QMUL-SDS              & Queen Mary University of London               & \citet{Zeng2021QMULSDSAS}       \\
    first\_1              & Ping An of China                               &    -                            \\
    SciKGAT               & Tsinghua University / Microsoft Research                     & \citet{Liu2020AdaptingOD}       \\
    bioBert for sciFact   &    -                                           &        -                        \\
    JC\_UKP               & TU Darmstadt                                  &    -                            \\
    sum\_rationale        &     UC Irvine                                          &    -                            \\
    base\_3\_4            &     -                                          &          -                      \\
    pasic\_scibert\_tfidf &       Ping An International Smart City                                        &      -                          \\
    \midrule
    \verisci              & Allen Institute for AI                        & \citet{Wadden2020FactOF}        \\
    \bottomrule
  \end{tabular}

  \caption{Systems that submitted to the \taskname leaderboard. Alongside system names are affiliations of the associated team (if applicable) and any papers associated with the submission (if applicable).}
  \label{tbl:teams}
\end{table*}

\section{Shared task description} \label{sec:task}

We briefly describe the dataset, task, and evaluation. Additional details on the data collection process and evaluation metrics can be found in \citet{Wadden2020FactOF}.

\paragraph{Dataset}
We use the \scifact dataset from \citet{Wadden2020FactOF}. \scifact consists of \numClaims claims with train, dev, and test splits of \numTrain, \numDev, and \numTest claims respectively. The full train and dev sets and test claims were publicly available six weeks prior to opening the submission portal; gold evidences and labels in the test set are not released publicly. Each claim is an assertion about a single biomedical entity or process, curated by a biomedical expert. These claims are verified by other biomedical experts against a corpus of \numPapers abstracts from peer-reviewed biomedical research articles. For each claim, relevant abstracts are those annotated with evidentiary sentences that \support or \refute the claim.

\paragraph{Task}
Given an input claim, the task is to (i) identify all abstracts in the corpus that are relevant to the claim, (ii) label the relation of each relevant abstract to the claim as either \supports or \refutes, and (iii) identify a set of evidentiary sentences (i.e. \emph{rationales}) to justify each label.

\paragraph{Evaluation}

Systems are evaluated by their F1 score for correctly identifying relevant abstracts. We report two evaluation metrics: \emph{abstract-level} F1 rewards a system for identifying and labeling abstracts correctly.  A system may predict up to three evidentiary sentences for each abstract. As long as these three sentences contain a gold rationale, the system is scored as correct; this is similar to the \fever score from \cite{Thorne2018FEVERAL}. In contrast, \emph{sentence-level} F1 rewards a system for identifying and labeling \emph{individual} evidentiary sentences correctly, similar to the ``conditional score'' introduced in \citet{Thorne2021EvidencebasedVF}. Unlike abstract-level evaluation, this metric penalizes models for over-predicting evidentiary sentences. In practice, we find that systems rank similarly in terms of sentence-level and abstract-level performance.

\paragraph{Model Submissions}

Submissions were made through a publicly-available online leaderboard\footnote{\url{https://leaderboard.allenai.org/scifact}}. To prevent overfitting on the test set, teams were permitted to make one submission per week. The leaderboard was available for seven weeks before final submissions were due.

\section{Overview of systems} \label{sec:systems}

Table \ref{tbl:teams} presents the submitted systems. As the online leaderboard is still accepting new submissions, we only include in this report the systems that were present by the shared task submission deadline. 

\subsection{Modeling approaches} \label{sec:components}

All systems for which model descriptions are available use a three-stage pipeline consisting of (1) retrieval of relevant abstracts, (2) selection of evidentiary sentences from retrieved abstracts, and (3) label prediction using the identified rationales. Many teams introduced improvements to these three components, which we summarize here.

\paragraph{Abstract retrieval}
Most systems rely on ``bag-of-words'' approaches such as TF-IDF or BM25 \cite{Robertson2009ThePR} to retrieve an initial set of candidate abstracts. In contrast, \paragraphJoint computes the distance between ``dense'' claim and abstract representations using BioSentVec \cite{Chen2019BioSentVecCS}.

Some systems further refine the initial set of retrieved abstracts using a neural model: \vertserini uses T5 \cite{Raffel2020ExploringTL} while \scikgat uses BERT \cite{Devlin2019BERTPO} to re-rank retrieved candidates. \qmul uses a BioBERT \cite{Lee2020BioBERTAP} text classifier for candidate filtering.

\paragraph{Evidence selection}

Systems take one of two approaches: \emph{single-sentence} systems predict whether a given sentence was evidence based on the claim and the sentence alone. \vertserini uses T5, \qmul uses BioBERT and \scikgat uses BERT to produce representations for each claim-sentence pair.

The \emph{full-document} systems encode the claim together with the entire abstract, and make predictions for each sentence based on this full-document encoding\footnote{Abstracts longer than 512 tokens are shortened by truncating long sentences}. The \paragraphJoint model interleaves \texttt{[SEP]} tokens between sentences, encodes the entire abstract using RoBERTa \cite{roberta-liu-2019}, and finally obtains sentence representations through self-attention pooling over words within each sentence. The Law \& Econ model treats rationale selection as a sequence tagging task, using a SciBERT \cite{beltagy-etal-2019-scibert} token-level tagger: any sentence with at least one predicted positive token is taken as evidence.

\paragraph{Label prediction}

Systems predict \support, \refute, or \nei (Not Enough Info) labels by concatenating the claim and all selected evidentiary sentences and feeding it through a neural three-way text classifier. Unless otherwise noted, systems tend to use the same model class for evidence selection and label prediction stages (e.g., for both pipeline stages, \vertserini use T5 and \paragraphJoint use RoBERTa).

The \scikgat system, which uses BERT for evidence selection, switches to using a kernel graph attention network \cite{Liu2020FinegrainedFV} for aggregating sentences for label prediction.  We note the \paragraphJoint team report experimenting with a similar approach but opt not to use it in their final system due to lack of positive results \cite{Li2021APM}.

The \qmul system switches from BioBERT to RoBERTa for the label prediction stage.  Furthermore, the model improves classification performance using a two-stage approach -- first classifying abstracts as either Containing Evidence or Not, and then classifying evidence-containing abstracts as Supported or Not.

\subsection{Model training techniques}  \label{sec:model_training}

We summarize several helpful techniques that teams used to improve model training when developing their systems.

\paragraph{Transfer learning}  \vertserini, Law \& Econ,  \paragraphJoint, and \scikgat models were first trained on the FEVER dataset \cite{Thorne2018FEVERAL} before training on \scifact. In contrast, \qmul was only trained on \scifact. \scikgat performed additional language model pretraining on data from the CORD-19 dataset \cite{Wang2020CORD19TC}.

\paragraph{Negative sampling}

Teams devised a number of strategies to expose the model to ``negative samples'' -- abstracts that do \emph{not} contain evidence relevant to a given claim. The \vertserini team used non-evidence sentences from relevant abstracts as negative samples. 

The \paragraphJoint team used irrelevant abstracts with high lexical similarity to ``gold'' relevant abstracts as additional negatives. Their system was trained using 12 negative abstracts per positive abstract, allowing it to maintain high precision while retrieving a larger number of candidate abstracts for each claim. This system also used \emph{scheduled sampling}, in which the label predictor is given gold rationales early in training and gradually transitions to using predicted rationales; this was found to increase model robustness.

Along similar lines, the \qmul team trained their evidence selection and label prediction components with retrieved abstracts rather than just using gold abstracts.

\paragraph{Dev set usage}

Both \vertserini and \paragraphJoint teams found it beneficial to perform initial hyperparameter selection on the dev set, and then train a final model on the train and dev sets combined. This is unsurprising given the moderate size of the \scifact training set.

\section{Results} \label{sec:results}



\begin{table*}[t]
  \footnotesize
  \setlength{\tabcolsep}{0.45em}
  \centering

  \begin{tabularx}{0.9 \linewidth}{*{1}{l} *{3}{Y} *{1}{s} *{3}{Y}}

    \toprule

    & \multicolumn{3}{c}{\textbf{Sentence-level}} & &  \multicolumn{3}{c}{\textbf{Abstract-level}} \\

    & \multicolumn{3}{c}{\textbf{Selection+Label}} & & \multicolumn{3}{c}{\textbf{Label+Rationale}}  \\

    Submission & P & R & F1 & &  P & R & F1 \\

    \midrule

    VerT5erini (Neural) (Train+Dev) & 60.59 & 66.49 & \textbf{63.40} & & 62.85 & 71.62 & 66.95          \\
ParagraphJoint                  & 68.94 & 54.59 & 60.94          & & 73.66 & 61.71 & \textbf{67.16} \\
VerT5erini (Neural)             & 60.00 & 57.57 & 58.76          & & 61.47 & 63.96 & 62.69          \\
Law \& Econ                     & 56.63 & 55.41 & 56.01          & & 62.80 & 58.56 & 60.61          \\
VerT5erini (BM25)               & 58.33 & 52.97 & 55.52          & & 60.28 & 58.11 & 59.17          \\
QMUL-SDS                        & 66.17 & 47.57 & 55.35          & & 72.97 & 48.65 & 58.38          \\
first\_1                        & 65.06 & 47.30 & 54.77          & & 65.36 & 52.70 & 58.35          \\
SciKGAT                         & 61.15 & 42.97 & 50.48          & & 76.09 & 47.30 & 58.33          \\
bioBert for sciFact             & 54.87 & 45.68 & 49.85          & & 58.29 & 49.10 & 53.30          \\
JC\_UKP                         & 40.69 & 41.35 & 41.02          & & 50.24 & 47.75 & 48.96          \\
sum\_rationale                  & 34.45 & 47.30 & 39.86          & & 42.38 & 51.35 & 46.44          \\
sum\_rationale                  & 28.45 & 52.97 & 37.02          & & 35.34 & 55.41 & 43.16          \\
base\_3\_4                      & 5.87  & 55.95 & 10.63          & & 9.54  & 50.45 & 16.05          \\
pasic\_scibert\_tfidf           & 7.95  & 14.32 & 10.22          & & 10.11 & 16.22 & 12.46          \\
\midrule
\verisci                         & 38.56 & 40.54 & 39.53          & & 46.61 & 46.40 & 46.50          \\
\verisci Zero-Shot               & 23.71 & 31.08 & 26.90          & & 42.26 & 31.98 & 36.41          \\

    \bottomrule

  \end{tabularx}

  \caption{System performance on the \taskname shared task. Systems are ordered by sentence-level F1. For \vertserini, (Neural) indicates that a neural re-ranker was used for retrieval, and (Train+Dev) indicates that the model was trained on both the train and dev sets as described in \S \ref{sec:model_training}.}
  \label{tbl:main_results}

\end{table*}

Table \ref{tbl:main_results} presents performance of all submissions on the public leaderboard during the seven week shared task period. \vertserini performs best on sentence-level evaluation, achieving an improvement of +23.87 F1 (+60.4\%) for sentence-level evaluation relative to the \verisci baseline. \paragraphJoint performs best on abstract-level evaluation, improving over the \verisci by +20.66 F1 (+44.4\%). Using the dev set as additional training data provides a substantial boost; this strategy alone improves \vertserini performance by roughly +5 F1. While \vertserini achieves higher recall, \paragraphJoint has higher precision; this is likely because \paragraphJoint was exposed to more negative samples during training.

\section{Insights}
\label{sec:insights}

A number of participating teams reported performance on individual pipeline components in publications associated with their systems (see Table~\ref{tbl:teams}). Based on their reports and discussions with shared task participants, we highlight findings on several modeling decisions that have had significant impact on results in the shared task.\footnote{Metrics reported in this section are self-reported by participants and have not been verified by the task organizers. Furthermore, metrics reported in this section are not directly comparable to those in Table~\ref{tbl:main_results} as the main results evaluate end-to-end system performance while the metrics reported here are ablations with respect to a pipeline component.}

\subsection*{Neural refinement of abstract candidates} 

The original \verisci model uses TF-IDF to retrieve the top $k=3$ documents for each claim, but shared task participants have demonstrated substantial improvement over ``bag-of-words'' retrievers using neural refinement. For instance, the \vertserini neural re-ranker improves Recall@3 from the TF-IDF score of 69.4 to 86.6, a +24.8\% increase. The \qmul system, which uses a binary classifier to filter for relevant abstracts, achieves an F1 of 74.2, while TF-IDF with a Top-3 strategy for identifying relevant abstracts only gets 26.3 F1. 

Can we simply replace the bag-of-words component with retrieval based on pre-trained dense representations of?  Not yet.  The \paragraphJoint team, which entirely replaced bag-of-words with dense embeddings obtained using BioSentVec \cite{Chen2019BioSentVecCS}, showed that abstract retrieval performance is slightly worse than simple TF-IDF, with a Recall@3 of 67.0. The \qmul team also informally reported negative results experimenting with pre-trained DPR \cite{karpukhin-etal-2020-dense}, another dense retrieval technique. These shared task findings comparing bag-of-words, neural refinement, and dense-only retrieval on \scifact abstract retrieval have also been demonstrated in other work (see Table 2 in \citet{thakur-beir-2021}). Adapting pre-trained dense retrieval models to \scifact remains an interesting area for future work.


\subsection*{Surrounding context for evidence selection} 

\paragraphJoint and Law \& Econ demonstrate positive results when incorporating full-abstract context during evidence selection, compared to independent assessment of each sentence. In an ablative analysis using \emph{oracle} abstracts, the \paragraphJoint team reports that incorporating full-abstract context increases evidence selection performance from 65.3 F1 to between 68.8 and 71.7 F1, relative improvements of +5.4\% to +9.8\%.

\subsection*{Reliance on larger models} These shared task results demonstrate the unreasonable effectiveness of simply using larger models to improve performance. In particular, \vertserini achieves substantial improvements while preserving much of the model architecture from the original \verisci baseline, but swapping in the larger T5 model. In an ablative experiment using \emph{gold} evidence, the \vertserini team reports that a label predictor that uses T5-3B achieves a performance of 88.2 F1, compared to 80.2 F1 when using RoBERTa-large, a +10.0\% relative improvement.  They observe similar improvements on evidence selection as well. In fact, \citet{Wadden2020FactOF} even reports higher performance by RoBERTa-large than SciBERT, where the former has 3 times more parameters, despite the latter being trained on in-domain data.

Still, \paragraphJoint demonstrates a surprising result in achieving performance competitive to \vertserini using RoBERTa-large (\numParamsRobertaLarge parameters, $\approx$10\% of T5-3B size), showing that other modeling strategies (e.g. negative sampling) can have significantly benefit system performance while keeping computational burden low.  First, this result suggests the need for an improved evaluation setup for the \taskname task that accounts for model weight classes for submissions and making comparisons.  Second, we postulate the need for a follow-up study investigating whether specific modeling strategies employed by smaller models would still translate to significant improvements when using larger models like T5.






\section{Conclusion} \label{sec:conclusion}

The \taskname shared task on scientific claim verification received 14 submissions from 11 teams. The submitted systems achieved substantial improvements on all three stages of the scientific claim verification pipeline -- abstract retrieval, evidentiary sentence identification, and label prediction -- improving on the previous state-of-the-art by 60\% and 44\% on the two primary evaluation metrics. We've surveyed the various approaches taken by participants and discussed several findings that have emerged from this shared task.  Looking forward, the strong performance of systems on \taskname suggests it may be time to explore other more ambitious challenges on the path toward building real-world scientific claim verification systems. For instance, future work could study retrieval from a much larger scientific corpus, approaches to consolidate evidence from multiple documents while taking into account the strength and credibility of different evidence sources, or techniques for building efficient systems that can approximate performance of the top submissions at a fraction of the cost.


\section*{Acknowledgements} 
We thank all the teams for their participation in the \taskname shared task. We are especially grateful to Ronak Pradeep and Xueguang Ma (\vertserini), Xiangci Li (\paragraphJoint), Dominik Stammbach (Law \& Econ), Xia Zeng (\qmul), and Zhenghao Liu and Chenyan Xiong (\scikgat) for their helpfulness in providing details about their systems and for participating in workshop presentations and discussions.  We also thank the organizing committee of the SDP 2021 workshop for hosting the \taskname shared task and for their feedback and assistance in helping us organize this event. We finally thank Iz Beltagy, Arman Cohan, Hanna Hajishirzi, and Lucy Lu Wang from AI2 for their help with reviewing, shared task insights and feedback in writing this report.

\bibliography{refs}

\begin{thebibliography}{19}
\expandafter\ifx\csname natexlab\endcsname\relax\def\natexlab#1{#1}\fi

\bibitem[{Beltagy et~al.(2019)Beltagy, Lo, and
  Cohan}]{beltagy-etal-2019-scibert}
Iz~Beltagy, Kyle Lo, and Arman Cohan. 2019.
\newblock \href {https://doi.org/10.18653/v1/D19-1371} {{S}ci{BERT}: A
  pretrained language model for scientific text}.
\newblock In \emph{EMNLP}.

\bibitem[{Chen et~al.(2019)Chen, Peng, and Lu}]{Chen2019BioSentVecCS}
Q.~Chen, Y.~Peng, and Z.~Lu. 2019.
\newblock \href {https://doi.org/10.1109/ICHI.2019.8904728} {{B}io{S}ent{V}ec:
  creating sentence embeddings for biomedical texts}.
\newblock In \emph{ICHI}.

\bibitem[{Devlin et~al.(2019)Devlin, Chang, Lee, and
  Toutanova}]{Devlin2019BERTPO}
Jacob Devlin, Ming-Wei Chang, Kenton Lee, and Kristina Toutanova. 2019.
\newblock \href {https://aclanthology.org/N19-1423} {{BERT}: Pre-training of
  deep bidirectional transformers for language understanding}.
\newblock In \emph{NAACL}.

\bibitem[{Karpukhin et~al.(2020)Karpukhin, Oguz, Min, Lewis, Wu, Edunov, Chen,
  and Yih}]{karpukhin-etal-2020-dense}
Vladimir Karpukhin, Barlas Oguz, Sewon Min, Patrick Lewis, Ledell Wu, Sergey
  Edunov, Danqi Chen, and Wen-tau Yih. 2020.
\newblock \href {https://aclanthology.org/2020.emnlp-main.550} {Dense passage
  retrieval for open-domain question answering}.
\newblock In \emph{EMNLP}.

\bibitem[{Lee et~al.(2020)Lee, Yoon, Kim, Kim, Kim, So, and
  Kang}]{Lee2020BioBERTAP}
Jinhyuk Lee, Wonjin Yoon, Sungdong Kim, Donghyeon Kim, Sunkyu Kim, Chan~Ho So,
  and Jaewoo Kang. 2020.
\newblock \href
  {https://academic.oup.com/bioinformatics/article/36/4/1234/5566506}
  {{BioBERT}: a pre-trained biomedical language representation model for
  biomedical text mining}.
\newblock \emph{Bioinformatics}.

\bibitem[{Li et~al.(2021)Li, Burns, and Peng}]{Li2021APM}
Xiangci Li, G.~Burns, and Nanyun Peng. 2021.
\newblock \href {http://ceur-ws.org/Vol-2831/paper8.pdf} {A paragraph-level
  multi-task learning model for scientific fact-verification}.
\newblock In \emph{Proceedings of the Workshop on Scientific Document
  Understanding}. AAAI.

\bibitem[{Liu et~al.(2019)Liu, Ott, Goyal, Du, Joshi, Chen, Levy, Lewis,
  Zettlemoyer, and Stoyanov}]{roberta-liu-2019}
Yinhan Liu, Myle Ott, Naman Goyal, Jingfei Du, Mandar Joshi, Danqi Chen, Omer
  Levy, Mike Lewis, Luke Zettlemoyer, and Veselin Stoyanov. 2019.
\newblock \href {http://arxiv.org/abs/1907.11692} {Roberta: {A} robustly
  optimized {BERT} pretraining approach}.
\newblock \emph{arXiv}, 1907.11692.

\bibitem[{Liu et~al.(2020{\natexlab{a}})Liu, Xiong, Dai, Sun, Sun, and
  Liu}]{Liu2020AdaptingOD}
Zhenghao Liu, Chenyan Xiong, Zhuyun Dai, Si~Sun, Maosong Sun, and Zhiyuan Liu.
  2020{\natexlab{a}}.
\newblock \href {https://doi.org/10.18653/v1/2020.findings-emnlp.216} {Adapting
  open domain fact extraction and verification to {COVID}-{FACT} through
  in-domain language modeling}.
\newblock In \emph{Findings of EMNLP}.

\bibitem[{Liu et~al.(2020{\natexlab{b}})Liu, Xiong, Sun, and
  Liu}]{Liu2020FinegrainedFV}
Zhenghao Liu, Chenyan Xiong, Maosong Sun, and Zhiyuan Liu. 2020{\natexlab{b}}.
\newblock \href {https://doi.org/10.18653/v1/2020.acl-main.655} {Fine-grained
  fact verification with kernel graph attention network}.
\newblock In \emph{ACL}.

\bibitem[{Pradeep et~al.(2021)Pradeep, Ma, Nogueira, and
  Lin}]{Pradeep2020ScientificCV}
Ronak Pradeep, Xueguang Ma, Rodrigo Nogueira, and Jimmy Lin. 2021.
\newblock \href {https://www.aclweb.org/anthology/2021.louhi-1.11} {Scientific
  claim verification with {V}er{T}5erini}.
\newblock In \emph{Proceedings of the 12th International Workshop on Health
  Text Mining and Information Analysis}. ACL.

\bibitem[{Raffel et~al.(2020)Raffel, Shazeer, Roberts, Lee, Narang, Matena,
  Zhou, Li, and Liu}]{Raffel2020ExploringTL}
Colin Raffel, Noam Shazeer, Adam Roberts, Katherine Lee, Sharan Narang, Michael
  Matena, Yanqi Zhou, Wei Li, and Peter~J. Liu. 2020.
\newblock \href {http://jmlr.org/papers/v21/20-074.html} {Exploring the limits
  of transfer learning with a unified text-to-text transformer}.
\newblock \emph{Journal of Machine Learning Research}, 21(140):1--67.

\bibitem[{Robertson and Zaragoza(2009)}]{Robertson2009ThePR}
Stephen Robertson and Hugo Zaragoza. 2009.
\newblock \href {https://doi.org/10.1561/1500000019} {The probabilistic
  relevance framework: {BM25} and beyond}.
\newblock \emph{Foundations and Trends in Information Retrieval},
  3(4):333–389.

\bibitem[{Stammbach and Ash(2020)}]{Stammbach2020eFEVEREA}
Dominik Stammbach and Elliott Ash. 2020.
\newblock \href
  {https://truthandtrustonline.com/wp-content/uploads/2020/10/TTO04.pdf}
  {e-{FEVER}: Explanations and summaries for automated fact checking}.
\newblock In \emph{TTO}.

\bibitem[{Thakur et~al.(2021)Thakur, Reimers, R{\"{u}}ckl{\'{e}}, Srivastava,
  and Gurevych}]{thakur-beir-2021}
Nandan Thakur, Nils Reimers, Andreas R{\"{u}}ckl{\'{e}}, Abhishek Srivastava,
  and Iryna Gurevych. 2021.
\newblock \href {https://arxiv.org/abs/2104.08663} {{BEIR:} {A} heterogenous
  benchmark for zero-shot evaluation of information retrieval models}.
\newblock \emph{arXiv}, 2104.08663.

\bibitem[{Thorne et~al.(2021)Thorne, Glockner, Vallejo, Vlachos, and
  Gurevych}]{Thorne2021EvidencebasedVF}
James Thorne, Max Glockner, Gisela Vallejo, Andreas Vlachos, and Iryna
  Gurevych. 2021.
\newblock \href {https://arxiv.org/abs/2104.00640} {Evidence-based verification
  for real world information needs}.
\newblock \emph{arXiv}, 2104.00640.

\bibitem[{Thorne et~al.(2018)Thorne, Vlachos, Christodoulopoulos, and
  Mittal}]{Thorne2018FEVERAL}
James Thorne, Andreas Vlachos, Christos Christodoulopoulos, and Arpit Mittal.
  2018.
\newblock \href {https://doi.org/10.18653/v1/N18-1074} {{FEVER}: a large-scale
  dataset for fact extraction and {VER}ification}.
\newblock In \emph{NAACL}.

\bibitem[{Wadden et~al.(2020)Wadden, Lin, Lo, Wang, van Zuylen, Cohan, and
  Hajishirzi}]{Wadden2020FactOF}
David Wadden, Shanchuan Lin, Kyle Lo, Lucy~Lu Wang, Madeleine van Zuylen, Arman
  Cohan, and Hannaneh Hajishirzi. 2020.
\newblock \href {https://doi.org/10.18653/v1/2020.emnlp-main.609} {Fact or
  fiction: Verifying scientific claims}.
\newblock In \emph{EMNLP}.

\bibitem[{Wang et~al.(2020)Wang, Lo, Chandrasekhar, Reas, Yang, Burdick, Eide,
  Funk, Katsis, Kinney, Li, Liu, Merrill, Mooney, Murdick, Rishi, Sheehan,
  Shen, Stilson, Wade, Wang, Wang, Wilhelm, Xie, Raymond, Weld, Etzioni, and
  Kohlmeier}]{Wang2020CORD19TC}
Lucy~Lu Wang, Kyle Lo, Yoganand Chandrasekhar, Russell Reas, Jiangjiang Yang,
  Doug Burdick, Darrin Eide, Kathryn Funk, Yannis Katsis, Rodney~Michael
  Kinney, Yunyao Li, Ziyang Liu, William Merrill, Paul Mooney, Dewey~A.
  Murdick, Devvret Rishi, Jerry Sheehan, Zhihong Shen, Brandon Stilson, Alex~D.
  Wade, Kuansan Wang, Nancy Xin~Ru Wang, Christopher Wilhelm, Boya Xie,
  Douglas~M. Raymond, Daniel~S. Weld, Oren Etzioni, and Sebastian Kohlmeier.
  2020.
\newblock \href {https://aclanthology.org/2020.nlpcovid19-acl.1} {{CORD-19}:
  The {COVID-19} open research dataset}.
\newblock In \emph{Proceedings of the 1st Workshop on {NLP} for {COVID-19}}.
  ACL.

\bibitem[{Zeng and Zubiaga(2021)}]{Zeng2021QMULSDSAS}
Xia Zeng and Arkaitz Zubiaga. 2021.
\newblock \href {https://doi.org/10.18653/v1/2021.sdp-1.15} {{QMUL}-{SDS} at
  {SCIVER}: Step-by-step binary classification for scientific claim
  verification}.
\newblock In \emph{Proceedings of the Second Workshop on Scholarly Document
  Processing}. ACL.

\end{thebibliography}
\bibliographystyle{acl_natbib}

\appendix

\end{document}